%% file: main.tex
\documentclass[letterpaper, 10 pt, journal, twoside]{IEEEtran}  

\IEEEoverridecommandlockouts                              

\usepackage{amsmath} 
\usepackage{amssymb}  
\usepackage[colorinlistoftodos]{todonotes}
\usepackage{hyperref}
\usepackage{multirow}
\usepackage{booktabs}
\usepackage[colorinlistoftodos]{todonotes}
\usepackage{siunitx}
\usepackage{url}
\usepackage{mathtools}
\usepackage{subcaption}
\usepackage{bm}  
\usepackage{soul}

\newcommand\copyrighttext{%
  \footnotesize \textcopyright 2019 IEEE.  Personal use of this material is permitted.  Permission from IEEE must be obtained for all other uses, in any current or future media, including reprinting/republishing this material for advertising or promotional purposes, creating new collective works, for resale or redistribution to servers or lists, or reuse of any copyrighted component of this work in other works.
  DOI: \href{<http://tex.stackexchange.com>}{https://doi.org/10.1109/LRA.2019.2927955}}
\newcommand\copyrightnotice{%
\begin{tikzpicture}[remember picture,overlay]
\node[anchor=south,yshift=4pt] at (current page.south) {\fbox{\parbox{\dimexpr\textwidth-\fboxsep-\fboxrule\relax}{\copyrighttext}}};
\end{tikzpicture}%
}

\providecommand{\keywords}[1]{\textbf{\textit{Index terms---}} #1}

\usepackage{accents}

\input{mathdef}

\title{Whole-Body MPC for a Dynamically Stable Mobile Manipulator}

\author{Maria Vittoria Minniti, Farbod Farshidian, Ruben Grandia, and Marco Hutter%
\thanks{Manuscript received: February, 24, 2019; Revised May, 23, 2019; Accepted June, 23, 2019.}
\thanks{This paper was recommended for publication by Editor Tamim Asfour upon evaluation of the Associate Editor and Reviewers' comments.
This work was supported in part by the Swiss National Science Foundation through the National Centre of Competence in Research Robotics (NCCR Robotics), in part by the Swiss National Science Foundation through the National Centre of Competence in Digital Fabrication (NCCR dfab), in part by Intel Network on Intelligent Systems, and in part by the European Union’s Horizon 2020 research and innovation programme under grant agreement No 780883.)} 
\thanks{All authors are with Robotic Systems Lab, ETH Zurich, Zurich 8092, Switzerland {\tt\footnotesize \{mminniti, farbodf, rgrandia, mahutter\}@ethz.ch}}%
\thanks{Digital Object Identifier (DOI): see top of this page.}
}

\begin{document}

\maketitle
\copyrightnotice

\begin{abstract}
Autonomous mobile manipulation offers a dual advantage of mobility provided by a mobile platform and dexterity afforded by the manipulator.  
In this paper, we present a whole-body optimal control framework to jointly solve the problems of manipulation, balancing and interaction as one optimization problem for an inherently unstable robot. The optimization is performed using a Model Predictive Control (MPC) approach;
the optimal control problem is transcribed at the end-effector space, treating the position and orientation tasks in the MPC planner, and skillfully planning for end-effector contact forces. The proposed formulation evaluates how the control decisions aimed at end-effector tracking and environment interaction will affect the balance of the system in the future.
We showcase the advantages of the proposed MPC approach on the example of a ball-balancing robot with a robotic manipulator and validate our controller in hardware experiments for tasks such as end-effector pose tracking and door opening.

\begin{IEEEkeywords}
Mobile Manipulation, Optimization and Optimal Control
\end{IEEEkeywords}

\end{abstract}

\section{Introduction}

\IEEEPARstart{R}{obots} that are balancing on a ball, i.e. ballbots, are interesting machines to study control principles due to their static instability when balancing on a single point contact \cite{satici2017intrinsic}. This control problem becomes even more challenging if one such robot is equipped with a multi-degrees-of-freedom (DOF) manipulator that can move objects and interact with the environment. The underlying dynamics of such robotic system is highly non-linear and unstable, which requires planning and control approaches that consider the full system dynamics. 

While ballbot balancing has been extensively studied in literature (see Section \ref{sec:related_work}), the extension with manipulation skills has not been investigated. In real world applications, the only existing work tackling manipulators of low complexity on a ballbot \cite{nagarajan2012planning} has only looked into control structures where the tridimensional, non-linear dynamics of the robot was approximated in two decoupled planes.

In this work, we present a  whole-body motion planning and control framework for mobile manipulators where the platform has unstable dynamics. The proposed method actively corrects for dynamic instabilities, making the closed loop system \textit{dynamically stable}, while enabling manipulation and interaction.

\begin{figure}[t]
\setlength\belowcaptionskip{-3ex}
\centering
\includegraphics[width=0.45\columnwidth]{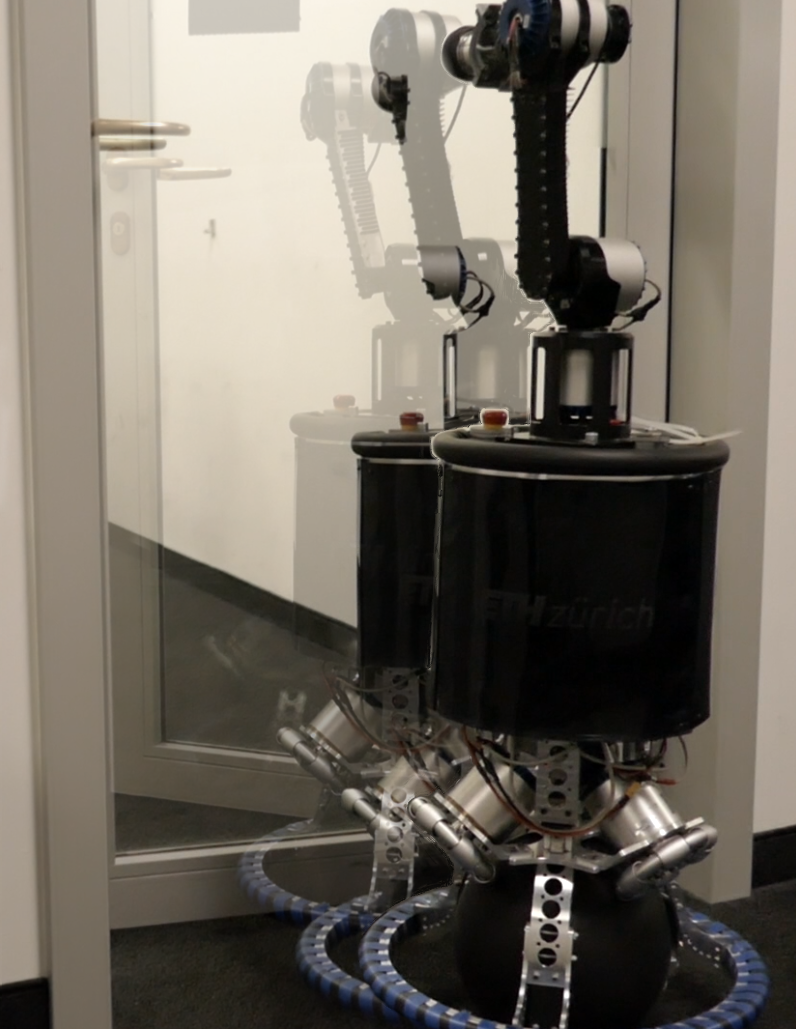}
\caption{Pushing door experiment with a dynamically stable mobile manipulator. A video showing the results is available at \url{https://youtu.be/mFNgbS3L3Ug}.
}
\label{fig:door_opening}
\end{figure}

Our contribution is demonstrated on a new design of Rezero, shown in Fig.~\ref{fig:door_opening}~and~\ref{fig:control_scheme}. This robot is a fully torque-controlled mobile manipulation system balancing on a single actuated ball, and it has a 3 DOF arm mounted on top. Many challenges arise in the control of an inherently unstable, non-minimum phase system when an arm is added. The additional inertia and the interaction between the dynamics of the arm and the mobile base result in a strong disturbance to the balancing task. End-effector control must be carefully designed in order not to compromise the balance of the platform.

The usual strategy to control mobile manipulators is to use a task space control scheme based on inverse dynamics \cite{siciliano2010robotics}. This approach achieves a satisfying performance when the manipulator under consideration is statically stable, such as a robotic arm on a fully actuated mobile base. However, when dealing with statically unstable mobile manipulators, which are underactuated, inverse dynamics control is more difficult to apply because of the presence of an unstable zero dynamics.

In the robotics literature, there are works that have treated the manipulation problem for systems balancing on unstable platforms (e.g., \cite{bellicoso2019}, \cite{mason2018mpc}, \cite{zafar2016whole}). However, current approaches do not take into account the full non-linear system dynamics in the planner. Therefore, they require more modeling assumptions and restrict the range of reachable tasks. Furthermore, by devicing a planner for balancing and a controller for end-effector tracking, they treat balancing and end-effector tasks separately in their control structure. These formulations introduce pre-defined levels of hierarchies in the controller definition, thus limiting the full expressiveness and capabilities of the approach.

In this paper, our main contribution consists in proposing an optimal control framework for treating both balancing and end-effector tasks in the same planning stage and over the same time horizon for a dynamically balancing mobile manipulator, while leveraging the full system dynamics. This allows the controller to take into account not only the tracking of the end-effector tasks, but also the stability over a time horizon. To further increase the robustness of our control structure, an MPC framework is used to adjust the control trajectory based on the most recent estimation of the system's state in a receding horizon fashion. 

Furthermore, we provide a description of how to incorporate a variety of tasks inside the optimal control problem. We show how it is possible to formulate the problem such that the system is capable of performing at the same time ball and end-effector tasks involving both motion control and interaction with the environment. The latter is obtained treating the end-effector contact forces as system inputs, and then adding a quadratic function of the force error inside the cost function. In order to improve the force planning task, inequality constraints are added to ensure that the friction limits are satisfied. 
The proposed approach is thoroughly evaluated in hardware experiments that resulted in the first demonstration of complex mobile manipulation with a ballbot system. 

\subsection{Related Work}
\label{sec:related_work}


Many ball-balancing robots have been built in the past years (\cite{lauwers2006dynamically}, \cite{kumagai2008development}, \cite{prieto2012monoball}, \cite{fong2009design}), although there are not many examples of such systems provided with manipulation capabilities. The first Ballbot system was introduced by Hollis et al. \cite{lauwers2006dynamically} at Carnegie Mellon University (CMU). 
The same group provided their robot with a pair of 2 DOF arms \cite{shomin2015sit}. However, the control problem employed in such works \cite{nagarajan2012planning} significantly differs from the one presented here. Indeed, the yaw motion for the base and arms of CMU Ballbot is not considered in the planner, which  therefore only uses a decoupled model of the system into two planes, limiting the range of possible tasks and the generalizability to more complex robots. 
 Furthermore, the balancing and manipulation tracking strategies are decoupled.
In \cite{asgari2015manipulation}, the manipulation problem is also addressed for a ballbot with an arm, where the model does not include the yaw DOF and the base is always controlled to be vertical, not allowing for simultaneous ball and arm tracking; results are only shown on a simulated model. As pointed out in \cite{satici2017intrinsic}, these modeling procedures, which model the ballbot motion into two planes, neglect the interaction effects of the full dynamics along the different planes. On the contrary, the controller presented here copes with the full non-linearity of the system and treats balancing and manipulation in a unified planner.

The first generation of Rezero \cite{fankhauser2010modeling} was controlled by interpolating LQR controllers at different operating points. Later, other control techniques were tested on this platform, such as a reinforcement learning method \cite{Farshidian2014} and an MPC framework \cite{neunert2016fast}. There are many novelties with respect to this previous work. First of all, the addition of the arm introduces a heavy disturbance to the stability of the system; therefore, in order to achieve manipulation tasks without compromising the balance of the platform, it is necessary to develop an approach that takes into account the coupling between the arm and the mobile base. In addition, unlike the old robot without the arm, this new upgraded version is provided with series elastic actuators. These provide a safe and compliant behavior, but have shown to be more difficult to control, due to bandwidth limitations and delays. To cope with this increased difficulty, the MPC controller has been modified according to the frequency shaping formulation presented in \cite{grandia2018frequency}. Most importantly, the focus of this work is how to tackle the simultaneous balancing and manipulation problem, and interaction with the environment. To this end, another novelty is the addition of end-effector motion control tasks in the optimal control problem formulation and the planning for contact forces. In this way the robot is allowed to balance and perform manipulation tasks on different end-effectors (ball and arm) at the same time. In terms of implementation, the current MPC planner is faster than the one presented in \cite{neunert2016fast}, can plan for a longer time horizon and allows for all the computations (state estimation, communication, MPC) to run on an on-board PC with moderate computational capabilities, even if the dimensionality of the system has increased. A constrained version of the optimal control problem is being used, while the previous one was unconstrained.
Unlike \cite{neunert2016fast}, where only a disturbance rejection and a go-to task were shown, here the proposed approach is validated with tests that show how the robot performs tasks  with different objectives at the same time, including tracking of arm and ball references and environment interaction.


So far, the most remarkable robot capable of performing dynamic coordination of locomotion, manipulation and balancing on a statically unstable platform is \textit{Handle} by Boston Dynamics \cite{bostondynamics_2017}, whose controller is not described in any publications. In \cite{bellicoso2019}, manipulation is performed with an arm on a quadrupedal robot. However, contact forces are considered as disturbances in the balancing planner. The performance or feasibility of the end-effector task is not taken into account while planning for balance.


In the MPC literature, a considerable effort has been put in showing real world applications on high dimensional systems \cite{koenemann2015whole}, but current approaches struggle with the problem of high computational cost. A way to address this issue is to plan center of mass trajectories for a reduced linear inverted pendulum model (\cite{mason2018mpc},\cite{mirjalili2018whole},\cite{zafar2018hierarchical}). 
These methods do not take into account the end-effector position and orientation tasks in the MPC planner. Hence, it is not possible to plan for the end effector motion and the balancing of the system over the same time horizon. Using a simplified model of the robot, the planner decisions do not consider the non-linearities and couplings of the system dynamics. Furthermore, since actuation torques are not considered during planning, these methods rely on more complex tracking controllers, adding another level of complexity to the control structure.

Compared to these works, since the outputs of the proposed planner are actuation torques and full state reference trajectories, it is possible to use a simple PD tracking controller. Furthermore, planning over end-effector position, orientation and force is included in the MPC formulation, which also takes care of the dynamic stability of the platform. 


\section{System}
\subsection{Physical System}
\label{sec:physical_system}

Rezero is a torque-controllable robot with a ball-balancing mobile base and a 3 DOF arm.  
The platform weighs about 25 kg. It is equipped with an onboard computer (Intel i7-7500U, 2.7 GHz, dual-core 64 bit) used for state estimation, motion control, and low-level actuators communication.  
Actuation is achieved by means of series elastic actuators \cite{bodie2016anypulator}. Three omni wheels are placed in a triangular layout between the main body and the ball, allowing the robot to move in any direction and to also rotate around its own axis.
The manipulator on top of the robot is a 3 DOF arm, where that the first joint rotates around the base vertical axis, and the other two rotate around an axis perpendicular to the first one. 

\subsection{System Modeling}
\label{sec:system_modeling}

In general, the motion of a mobile manipulator with an arm is characterized by $n_b + n_a$ degrees of freedom, where $n_b$ is the number of free variables of the mobile base, and $n_a$ is the number of degrees of freedom of the arm. 

\begin{figure}[t]
\setlength\belowcaptionskip{-3ex}
\vspace{-3ex}
\centering
\includegraphics[scale=0.25]{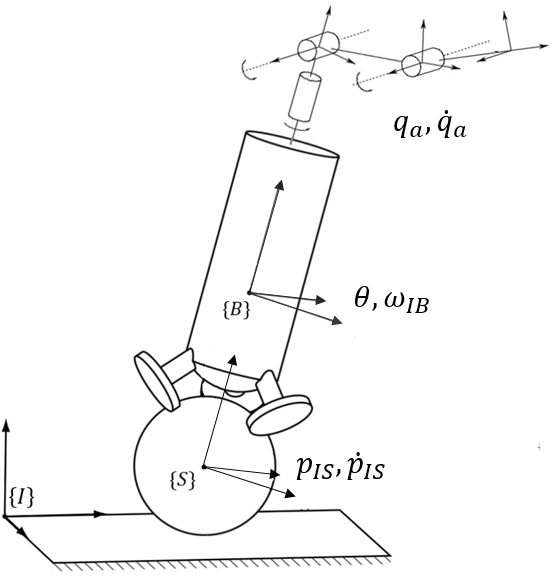}
\caption{Kinematic scheme of Rezero. $\{I\}, \{S\}, \{B\}$ are the inertial frame, the ball (sphere) frame and the base frame.}
\label{fig:ballbot_kinematics}
\end{figure}

We describe the motion of Rezero with a minimal set of generalized coordinates. These coordinates are the 2D position of the ball in the horizontal plane, the orientation of the base, and 3 joint angles of the arm as depicted in Fig.~\ref{fig:ballbot_kinematics}. We refer to the base as the main body of the robot, between the ball and the arm.
The system has in total 8 degrees of freedom ($n_b = 5, n_a = 3$). Rezero is equipped with 6 actuators ($n_{\tau}=6$), 3 for actuating the ball and 3 at each joint of the arm. Since the number of degrees of freedom of the robot is larger than the number of independent control inputs, the system is underactuated.

In Fig.~\ref{fig:ballbot_kinematics}, the frames and variables used for the kinematics of the ballbot are depicted. $\{I\}$ is the inertial reference frame, $\{B\}$ is a frame rigidly attached to the base body, and $\{S\}$ is a frame parallel to the base frame and located at the center of the ball.  

We introduce the vector of generalized coordinates and velocities $\vq, \vnu \in \mathbb{R}^{n_b + n_a}$,  given by:
\begin{equation}
\vq = \begin{bmatrix}
    \bm{p_{IS}} \\
    \vth \\
    \vq_a
    \end{bmatrix}, \quad \vnu = \begin{bmatrix}
    \dot \vp_{IS} \\ \vom_{IB} \\ \dot \vq_a
    \end{bmatrix},
\end{equation}  
where $\vp_{IS} \in \mathbb{R}^2$, $\vth \in \mathbb{R}^3$ and $\vq_a \in \mathbb{R}^{n_a}$ indicate the ball's position in the horizontal plane, a vector of ZYX Euler angles representing the orientation of the base, and the joint coordinates of the arm, respectively. $\vom_{IB} \in \mathbb{R}^3$ denotes the base angular velocity relative to the inertial frame.

Having chosen the vectors $\vq$ and  $\vnu$, we treat the robot as a dynamical system with state vector $\vx=(\vq, \vnu) \in \mathbb{R}^{2(n_b+n_a)}$
and input vector
$\vu=(\vtau, \vlambda_{ee}) \in \mathbb{R}^{n_{\tau}+n_{\lambda}}$, where $\vtau$ and $\vlambda_{ee}$ are the actuation torques and the forces exerted by the environment on the end-effector, which may include both linear forces and moments. 

Then, the state equations are given by $\dot \vx = \vf(\vx, \vu)$, where
\begin{equation}
\label{state_equations}
\small
    \vf(\vx, \vu) \coloneqq 
    \begin{bmatrix}
    \vT(\vq) \vnu \\
    \vM(\vq)^{-1}(-\vh(\vq, \vnu) + \vS^{T}(\vq)\vtau + {\vJ^T}_{ee}(\vq)\vlambda_{ee})
    \end{bmatrix}
    .
\end{equation}
Here, $\vT(\vq) \in \mathbb{R}^{n_b+n_a \times n_b+n_a}$ maps the vector of generalized velocities to the vector of generalized coordinate derivatives. $\vM(\vq)$, $\vh(\vq)$ and $\vJ_{ee}(\vq)$ are the system mass matrix, vector of non linear effects, and end-effector jacobian, respectively. The actuator selection matrix $\vS(\vq)$ depends on the generalized coordinates due to how the ball is actuated.

\section{Control Method}

\subsection{Optimal Control Problem}
\label{section:optimal_control_problem}
In this section, we describe the optimization problem and a general form of the employed cost function, which are the key points of our control approach.


Let $\vp_{IE} \in \mathbb{R}^3 $ be the position of the end-effector with respect to the inertial frame. We also define $\ve_o \in \mathbb{R}^3 $ to be an orientation error, which describes the deviation between a desired orientation and a measured one. Depending on the chosen parameterization for the end-effector orientation, there are multiple ways to express such an error. The reader can find additional details in \cite{siciliano2010robotics}, where all the different forms for the orientation error are exhaustively explained. For the sake of completeness, we briefly describe the quaternion distance measure used in this paper.

Let $\vPh_d = \{\eta_d, \vep_d\}$, $\vPh_e = \{\eta_e, \vep_e\}$ be unit quaternions representing the desired end-effector orientation and the measured one, respectively. If the desired end-effector frame and the measured one coincide, the equations
$
\vDe\vPh = \vPh_d \star \vPh_e^{-1} = \{1, \mathbf{0}\}
\label{quaternion_equality_condition}
$
must be satisfied, where the symbol $\star$ indicates the quaternion product operation. This yields the following form for the orientation error:
\begin{equation}
\ve_o = \eta_e \vep_d - \eta_d \vep_e - [\vep_d]^{\times}\vep_e
\end{equation}
where $[\cdot]^{\times}$ indicates the skew matrix operator.

Furthermore, we define $\vxi = ( \vv, \vom) \in \mathbb{R}^6$ as the twist describing the motion of a rigid body, which includes linear and angular velocity. We will use subscripts to indicate the rigid bodies to which these variables refer.

For every rigid body, our algorithm allows planning over pose, twist, and wrench. Therefore the set of optimization variables is $SE(3) \times \mathbb{R}^6 \times \mathbb{R}^6$. 
According to our modeling framework, the motion and force planning problem can be written in terms of the following minimization problem 

\begin{align}
    J 
    &= \min_{\vtau(\cdot), \vlambda_{ee}(\cdot)} \frac{1}{2}\int_0^T 
    \Big[ 
    \Vert \vp_{IE}^{d} - \vp_{IE} \Vert^2_{\vQ^{ee}_{pos}} 
    + \Vert \vp_{IS}^{d} - \vp_{IS} \Vert^2_{\vQ^s_{pos}} 
    \notag
    \\
    &\phantom{= \min_{u(\cdot)} \frac{1}{2}\int}
    + \Vert {\ve_o}_{IE} \Vert^2_{\vQ^{ee}_{or}}
    +  \Vert {\ve_o}_{IS} \Vert^2_{\vQ^{s}_{or}} 
    +\Vert \vxi_{IE}^d - \vxi_{IE} \Vert^2_{\vQ^{ee}_{vel}} 
    \notag
    \\
    &\phantom{= \min_{u(\cdot)} \frac{1}{2}\int}
    + \Vert \vxi_{IS}^d - \vxi_{IS} \Vert^2_{\vQ^{s}_{vel}}+ \Vert \vtau \Vert^2_{\vR_{\tau}}
    \notag
    \\
    &\phantom{= \min_{u(\cdot)} \frac{1}{2}\int}
    + \Vert \vlambda_{ee}^{d} -\vlambda_{ee} \Vert^2_{\vR_{\lambda}} 
    \Big] dt,
    \label{cost}
\end{align}
such that the system's equations of motion and correspondent initial conditions are satisfied
\begin{equation}
\label{equations_of_motion}
    \dot \vx = \vf(\vx, \vu), \quad  \vx(0) = \vx_0.
\end{equation}

along with the friction cone constraint for the linear force at the end-effector and torque limits:

\begin{align}
    &\sqrt{{\lambda_{ee}}_x^2 + {\lambda_{ee}}_y^2} \leq \mu {\lambda_{ee}}_z, \notag 
    \\
    \label{inequality_constraints}
    &\vtau_{min} \leq \vtau \leq \vtau_{max},
\end{align}

In \eqref{inequality_constraints}, $\mu$ is a friction coefficient; we assume that the contact force at the end-effector is expressed in a local frame such that the $z$ axis is normal to the contact surface. Inequality constraints are implemented according to a relaxed barrier function formulation, introduced in \cite{1905.06144}.

Each of the quadratic terms in the cost function corresponds to a task that the robot is required to perform, while satisfying the other tasks at the same time. In \eqref{cost},  $\vQ^{ee}_{pos}$, $\vQ^{ee}_{or}$, $\vQ^{ee}_{vel}$, $\vQ^s_{pos}$, $\vQ^{s}_{or}$, $\vQ^s_{vel}$ are semidefinite positive matrices which weight position, orientation and (linear and angular) velocity errors at the end-effector and ball frame, respectively. As introduced in section \ref{sec:system_modeling}, since in our robot model a frame parallel to the base frame is located at the center of the ball, it is equivalent to control the ball or base orientation and angular velocity.
$\vR_{\lambda}$ is a positive definite matrix for the end-effector wrench error. Finally, to penalize high torque commands, we also add a quadratic penalty on the torque inputs with the weight matrix $\vR_{\tau}$.  

The choice $\vQ^s_{pos} = \mathbf{0}$ gives the algorithm the freedom to plan a desired motion of the ball and the arm in order to track a desired end-effector pose. However, sometimes it may be useful to have $\mathbf{Q}^s_{pos} \neq \mathbf{0}$. This is required if the ball of the robot needs to execute specific trajectories while performing manipulation tasks.

Similarly, we may wish not to optimize over end-effector forces and solve a pure motion control problem. Since both $\vR_{\tau}, \vR_{\lambda}$ must be positive definite, there are two possibilities. One is the exclusion of force inputs. Otherwise, the same effect is achieved setting $\vlambda_{ee}^d = \mathbf{0}$.


For numerical reasons, it may be useful to add a regularization term to the cost function penalizing the arm joint velocities.

For completeness, in \eqref{cost}, we added a quadratic term with respect to ${\ve_o}_{IS}$. We clarify here that we always only penalize the yaw component $\theta_z$ of the base orientation, and let the controller optimize for $\theta_y$ and $\theta_x$ based on the desired task. Therefore, the balancing task for the robot is only specified through the last three components of the matrix $\vQ^s_{vel}$, which weigh the  base angular velocity error.

\subsection{SLQ-MPC Control}

The nonlinear optimal control problem defined in equations \eqref{cost}, \eqref{equations_of_motion} and \eqref{inequality_constraints} has been solved using the \textit{Sequential Linear Quadratic} (SLQ) algorithm in a model predictive control fashion; details on the employed solver and the MPC scheme can be found in \cite{farshidian2017efficient, Farshidian2017MPC}.

In summary, this algorithm designs a continuous controller for a nonlinear system. In addition, it has a linear computational complexity with respect to the optimization time-horizon.

In each iteration, the continuous-time SLQ algorithm forward integrates the system dynamics over the time horizon based on the last approximation of the optimal controller. 
Then, it linearizes the system dynamics and quadratizes the cost function over the state and input trajectories derived from the forward integration.

Finally, the algorithm solves a constrained, time-varying LQR problem based on the quadratic cost function and linearized system.
As a result, the updated control input has the form
\begin{equation}
    \label{eq:updated_control_input}
    \vu(t,\vx) = \vu_{ff}(t) + \alpha \, \vdelta\vu_{ff}(t) + \vK(t)(\vx(t) - \vx_{ff}(t)),
\end{equation}
where $\alpha$ is a line-search parameter used to update the feedforward law.

This algorithm runs a real-time iteration MPC scheme \cite{diehl2005}, where the solver is triggered by a new state measurement and uses a warm start. In such a scheme, the optimal control solver does not run until convergence: after the real-time iteration, it moves to the next MPC problem.

For the tracking part, we use the feedforward torques, positions and velocities $\vtau_{ff}$, $\vq_{ff}$, $\vnu_{ff}$, which are outputs of the SLQ-MPC algorithm. Since the solver is continuous time, the discretization is carried out by the ODE solver. During execution, the feedforward trajectory is linearly interpolated around the discretized points.

\subsection{Frequency-shaped cost function}
So far, we have assumed perfect tracking of the actuators torques. This is usually not true when dealing with series elastic actuators, such as those used during the experiments. To deal with this modeling mismatch, one possibility is to slightly modify the optimal control problem according to a frequency aware approach, which is described in \cite{grandia2018frequency}.
In particular, instead of considering directly the wheels torques $\vtau_w$ in the cost function, we use a filtered version of those. In frequency:

\begin{equation}
\label{filtered_wheels_torques}
{\hat \tau_w}_i(\omega) = r_i(\omega) {\tau_w}_i(\omega), \quad i = 1,2,3
\end{equation}
where $r_i(\omega)$ is a first-order filter designed based on the frequency response of the actuators. To implement this, we need to add a dynamical system corresponding to the filter dynamics in cascade with the robot's dynamical system, thus increasing the number of filter states from $16$ to $19$.

Such method allows to consider bandwidth limitations due to the presence of spring elements and has been successfully used to improve torque tracking of the ball actuators.

\subsection{Tracking Problem}
Depending on the complexity of the minimization problem, the frequency of the MPC planner can be less than the control loop frequency. Hence, we need to close a local feedback loop to improve the control performance. 

Therefore, it is useful to introduce a set of actuation positions and velocities, that we define as $\vq_{act}$, $\dot \vq_{act}$ and that are functions of both both $\vq$, $\vnu$. These, for the system under consideration, are the positions and velocities of the joints of the wheels and the arm. 

Then, actuation torques can be computed according to:
\begin{equation}
    \vtau_{act} = \vtau_{ff} + \vK_D ({\dot \vq_{act}}_{ff} - \dot \vq_{act}) + \vK_P ({\vq_{act}}_{ff} - \vq_{act}),
\label{tracking_controller_equation}
\end{equation}
where $\vK_P$, $\vK_D$ are two positive definite diagonal matrices.

The feedback terms in equation \eqref{tracking_controller_equation} are used to guarantee stability and represent a small contribution to the total torque. Indeed, the elements of $\vK_P$, $\vK_D$ are very small and, as it will be shown in the results section, the actuators behavior is mainly dictated by the feedforward term $\vtau_{ff}$.

\begin{figure}[t]
\setlength\belowcaptionskip{-3ex}
\centering
\includegraphics[scale=0.35]{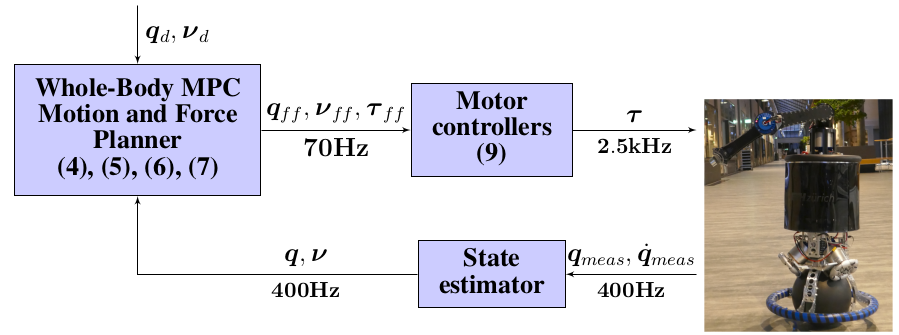}
\caption{Schematic representation of the controller structure. Numbers in parenthesis indicate the corresponding equations.
}
\label{fig:control_scheme}
\end{figure}

\subsection{Automatic Differentiation}
As other model-based optimization methods, the MPC planner proposed in this paper needs to compute the linearization of the system dynamics and of the cost.
Hence, an automatic differentiation tool (such as Auto-Diff) is required to compute derivatives.
The system dynamics includes the non-linear rigid-body equations of motion, while the cost is a function of the end-effector forward kinematics map; in order to be compatible with automatic differentiation, both have to be functions of a special Auto-Diff scalar type.

In \cite{neunert2016fast}, this problem was not addressed because the provided cost function was already quadratic with respect to the system state (there was no end-effector planning) and the equations of motion were stored symbolically, along with their derivatives. This procedure was already memory-consuming for the system without the arm, and it became inapplicable because of the more complex kinematic chain of the robot with the manipulator.

In this work, RobCoGen \cite{giftthaler2017automatic} was employed for this scope: an efficient code generation framework for modeling Rigid Body Dynamics, which offers the feature to provide rigid body kinematics and dynamics quantities in terms of the required scalar type.

\section{Results}

In this section, we validate the performance of our control algorithm. 
In each test, all computations run on the robot's onboard computer (refer to subsection~\ref{sec:physical_system}). 
The controller relies on a state estimator running at 400 Hz, which fuses odometry measurements from the motors' encoders and IMU information to get an estimate of the base pose, linear and angular velocity. The algorithm used for state estimation is based on \cite{bloesch2018two}. The MPC loop runs in a separate ROS node at a frequency of approximately 70 Hz, meaning that every $1/70$ s the solver optimizes the trajectory based on the latest state measurement. In all the experiments, the time horizon has been fixed to a value between 1s and 2s, that is the range under which the controller has been tested. The tracking controller is implemented inside the internal motor controllers and runs at 2.5 KHz. The structure of the controller is schematically described in Fig.~\ref{fig:control_scheme}.

To validate the performance of our proposed motion control approach, we have conducted four experiments. 
These experiments showcase different aspects of our proposed control approach introduced in section~\ref{section:optimal_control_problem}. 
The first experiment focuses on the ability of the controller to coordinate the motion of the base and the arm while the task is just defined at the arm's end-effector space. 
The second experiment demonstrates the disturbance rejection capability of the controller for two different tasks, the ball position control, and the end-effector's position control task.
The third experiment highlights the necessity of a whole-body control approach for Rezero. The experiment compares the behavior of the robot under two different control schemes: the proposed whole-body control approach and a method in which the base and the arm are controlled separately. 
Finally, the last experiment demonstrates how the control scheme skillfully incorporates the reference contact forces at the end-effector into the motion planning for a door opening task.     
All the results are also presented in the video attachment \footnote{\url{https://youtu.be/mFNgbS3L3Ug}} which shows the ability of the robot to perform different tasks.

\subsection{End-effector Reference Pose Tracking}

In this experiment, we focus on end-effector pose tracking while ensuring the balance of the robot. 
To this end, we set the cost weight for the ball pose and twist to zero. 
However, we penalize the end-effector pose error and add a small penalty on its twist. 
An important aspect of this experiment is that without specifying the base motion reference, the controller coordinates the motion of the base and the arm in an optimal way such that Rezero fulfills its end-effector task while dynamically balancing.

\begin{figure}[t]
\setlength\belowcaptionskip{-3ex}
\centering
\includegraphics[width=\columnwidth]{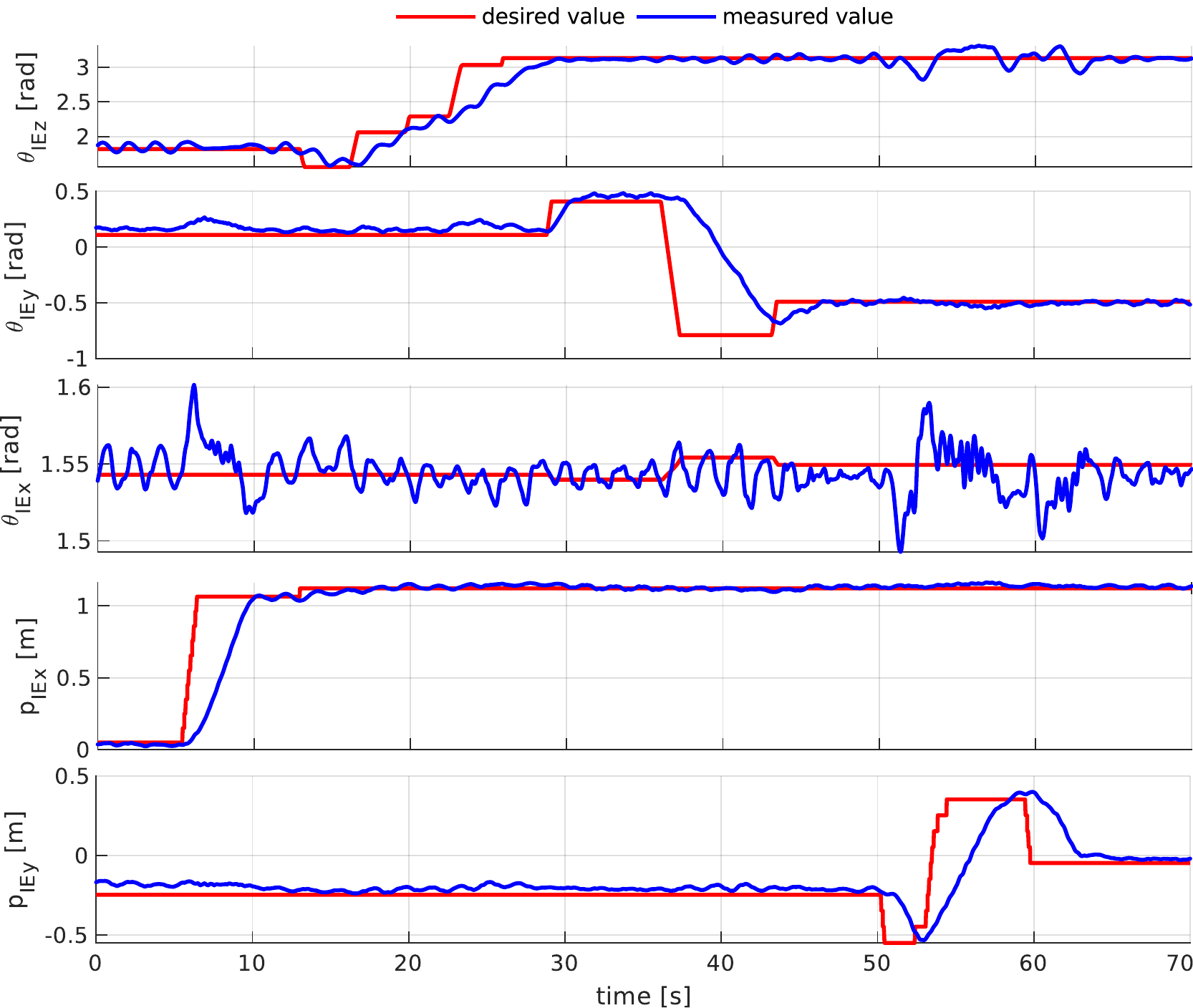}
\caption{Response of end-effector position ($\vp_{IE}$) and euler angles ($\vth_{IE}$) to various step references. Here we minimize desired end-effector position ($x$ and $y$) error, orientation error, base yaw error, joint velocities, and joint torques. The base roll and pitch angles and the ball position result from the minimization problem to track the desired end-effector pose.}
\label{fig:ee_reference_tracking}
\end{figure}

The tracking performance of the end-effector is demonstrated in Fig.~\ref{fig:ee_reference_tracking}, which is the robot's response to a sequence of step references for the end-effector's pose in the world frame. 
In these plots, the end-effector pose tracks the reference commands with an acceptable rise time response and steady-state error. 
The quality of the motion can be modified by setting higher weights on the pose tracking error and reducing the penalty over the end-effector's twist.  
In the attached video, you can observe how the base orientation is continuously adapting to the motion of the arm so that the platform remains stable. 

Note that due to the lack of degrees of freedom at the arm, in order to change the rolling along the normal axis of the end-effector ($x$ axis), the base of the robot should tilt. However, since the robot keeps its balance by regulating its tilt angles, the robot cannot track an arbitrary reference around this direction. This effect can be observed in the third plot of Fig.~\ref{fig:ee_reference_tracking}. This aspect highlights the capability of our optimal controller to make a trade-off between the tracking error minimization at the current instant and the balance of the system in the future.   

\subsection{Disturbance Rejection}

The primary purpose of this experiment is to demonstrate the disturbance rejection capability of our motion controller. To this end, we have designed two tasks. In the first task, we are only interested in controlling the ball position. Thus, the cost weight for the ball position is high. As for the arm, we only weigh the end-effector orientation error and we do not penalize end-effector position error. In the second task, we intend to control the position of the end-effector. Therefore, we place a higher weight for the end-effector's position error. 

\begin{figure}[t]
\centering
\includegraphics[width=\columnwidth]{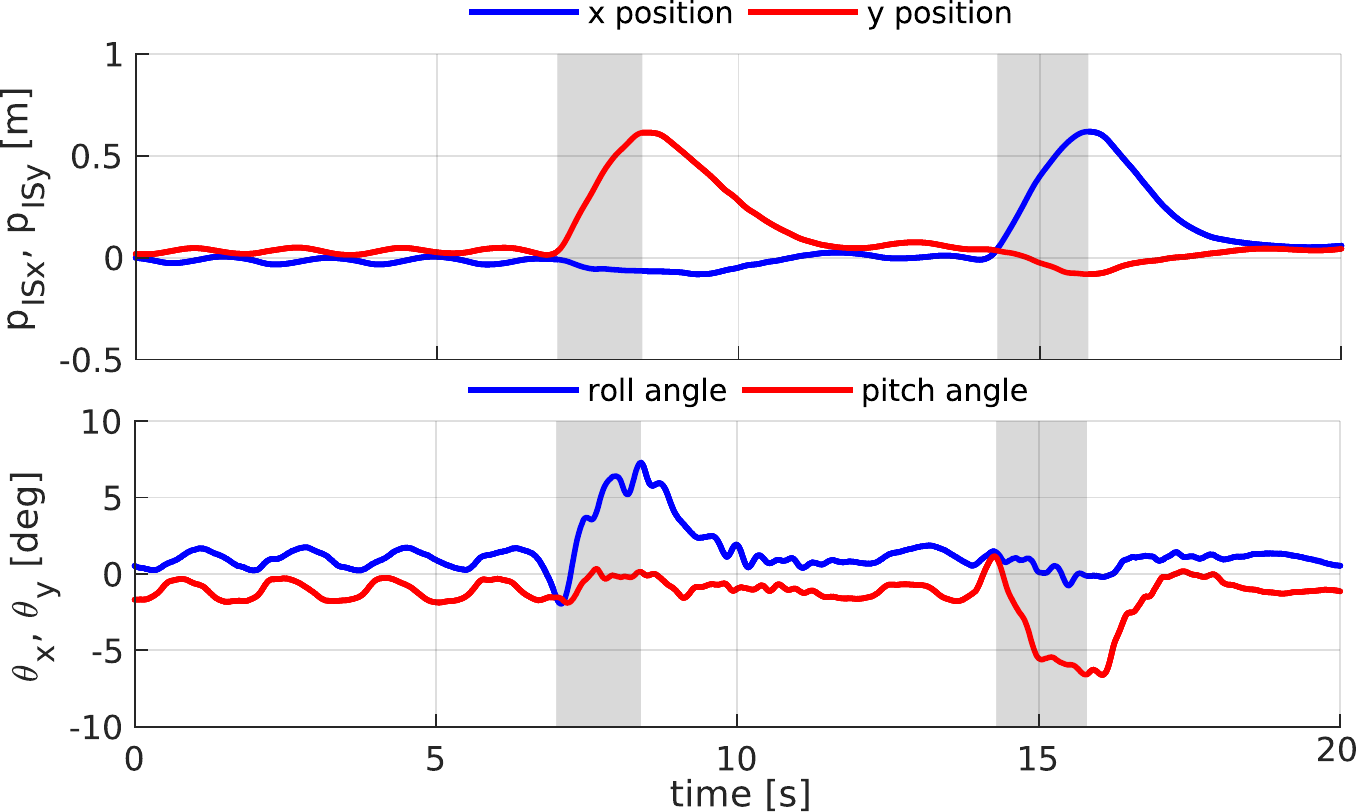}
\caption{Evolution of the ball position and roll and pitch angles during a disturbance rejection task. In the upper plot, we depict the trajectories of the ball position. The robot is perturbed twice during the experiment, which corresponds to the gray areas. In the lower plot we depict the motion of the base orientation (roll and pitch angles) during the experiment.}
\label{fig:ball_disturbance_rejection}
\end{figure}

For the first task experiment, we perturb the robot twice by pushing it away from its initial position. Our disturbances are applied successively along the $y$ and $x$ directions. 
The evolution of the ball position and roll and pitch angles is shown in Fig.~\ref{fig:ball_disturbance_rejection}.
The robot maintains its balance by tilting its base pose against perturbation direction (see the attached video). This is a natural behavior where the robot shifts its center of the mass against the perturbation.

In the attached video, we show the disturbance rejection behavior of the second task where we control the end-effector position. Similarly to when the controlled variable is the ball position, the robot reacts to the disturbance with the motion of its whole body. However, in this experiment, it uses its arm more prominently than before.  

\subsection{The Necessity of the Whole-body Control}

\begin{figure}[t]
\setlength\belowcaptionskip{-3ex}
\centering
    \begin{subfigure}[b]{0.115\textwidth}
        \centering\includegraphics[scale = 0.13]{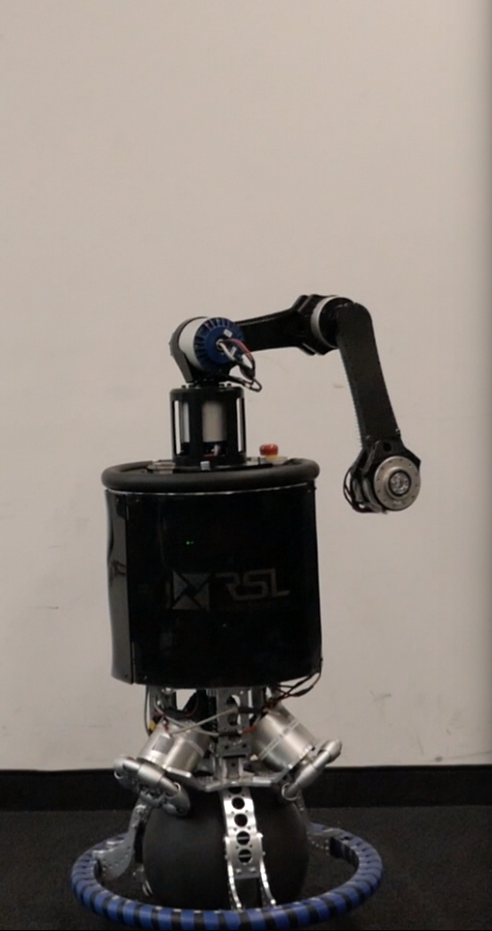}
    \end{subfigure}
    ~ 
    \begin{subfigure}[b]{0.127\textwidth}
        \centering\includegraphics[scale = 0.13]{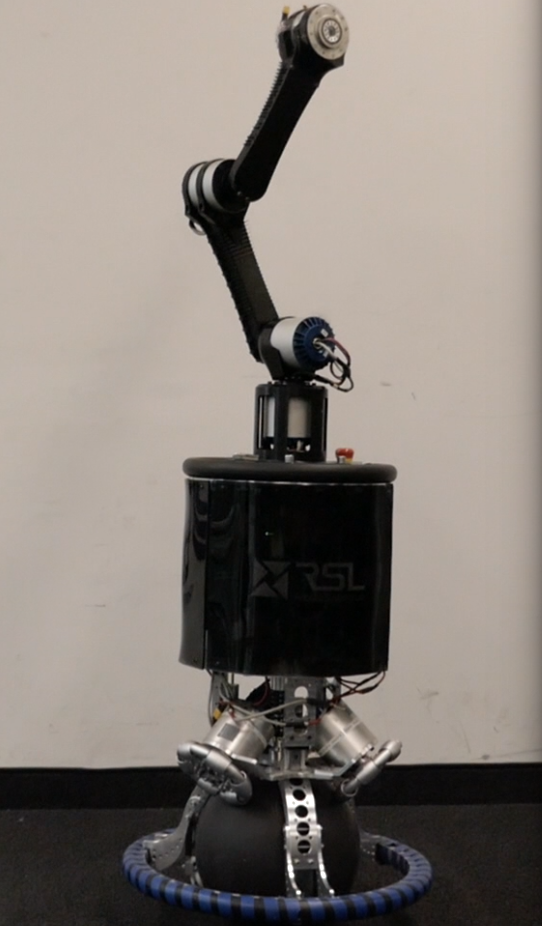}
    \end{subfigure}
    ~ 
    \begin{subfigure}[b]{0.11\textwidth}
        \centering\includegraphics[scale = 0.13]{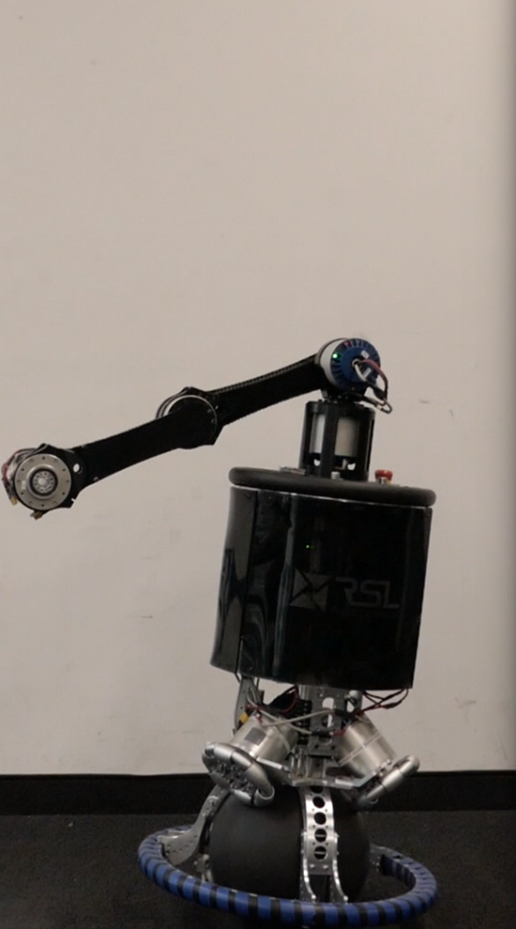}
    \end{subfigure}
    \caption{The robot is given a sequence of end-effector position commands, while the ball is controlled to remain in the same position. The base orientation adapts to the different arm configurations.}\label{_sequence}
\end{figure}

Here, we are interested in controlling the ball and the end-effector position at the same time.
As a showcase for this possibility, we give the end-effector a sequence of reference positions, while setting the initial position as a goal for the ball.
In Fig.~\ref{_sequence}, we show how the base adapts to the center of mass location due to the arm motion. Since the robot lower body is consistently heavier than the manipulator, we added weight to the end-effector to highlight the base adaptation.

\begin{figure}[t]
\centering
\setlength\belowcaptionskip{-3ex}
\includegraphics[width=0.9\columnwidth]{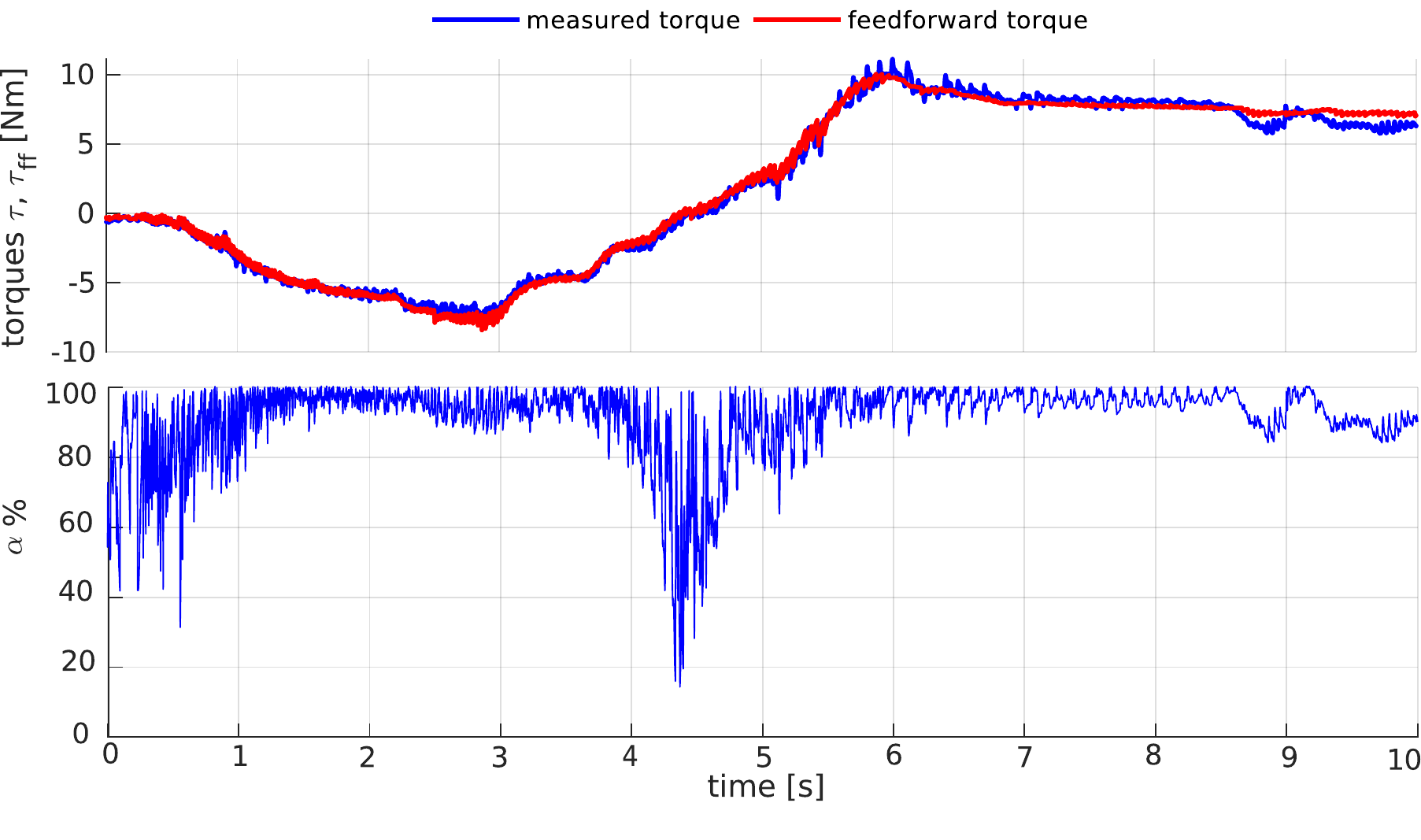}
\caption{Contribution of the feedforward torque to the measured torque for the second joint of the arm. In the upper plot we show the measured and feedforward torque, while in the lower plot we show the ratio $\alpha$ in percentage from equation \eqref{torques_ratio}. } 
\label{fig:feedforward_term_contribution}
\end{figure}

Furthermore, to manifest that the feedforward term represents the major contribution of the total torque in equation \eqref{tracking_controller_equation}, in Fig.~\ref{fig:feedforward_term_contribution}, we compare the trajectories of the measured torque and the feedforward torque for the second joint of the arm during the experiment. We also compute the ratio
\begin{equation}
\small
    \alpha \% \coloneqq 100 \frac{|\vtau_{ff}|}{|\vtau_{ff}| + |\vtau_{fb}|} ,
    \label{torques_ratio}
\end{equation}
where $\vtau_{fb}$ represents the feedback part, and plot it in percentage. 

As another example of both ball and end-effector position tracking, in the video attachment we show the behavior of the robot while we command the ball to move on a circular trajectory and the end-effector to remain in the same position.

To further highlight the importance of our whole-body control approach, in the attached video, we show the performance of a setup which has two separate modules for controlling the base and the arm.   
Here, we use a joint-level PID controller for the arm. This controller transforms the transcribed task at the end-effector to a joint level task using an inverse kinematics model. 
For the mobile base, we use a degraded version of our MPC method such that it only controls the base. 
This control only considers the arm at a nominal configuration. 
However, it does not know about the arm motion, and it only receives updated measurements of the base and the ball.  
Therefore, the base controller cannot account for changes in dynamics due to the arm motion.

Similar to the whole-body control experiment, we command the ball to keep a fixed position, while we move the arm's end-effector references. 
Contrary to the whole-body controller, the ball position does not remain fixed as the arm changes configuration (refer to the attached video). 
This experiment manifests the importance of a whole-body control approach for coordinated control of the base and the manipulator in dynamically stable platforms.

\subsection{End-effector Force Planning}

In this experiment we want the controller to plan a desired force trajectory at the end-effector in order to push and open a door (see Fig.~\ref{fig:door_opening}).
Therefore, we have not assigned any cost on the end-effector position and velocity. However, we give a penalty to the end-effector orientation and to the error between a desired linear force and a measured one.
Inequality constraints are added for the friction cone (friction coefficient is set to $\mu = 0.7$) and for torque limits.

\begin{figure*}
\setlength\belowcaptionskip{-4ex}
  \begin{subfigure}{8.5cm}
    \centering\includegraphics[width=0.9\columnwidth]{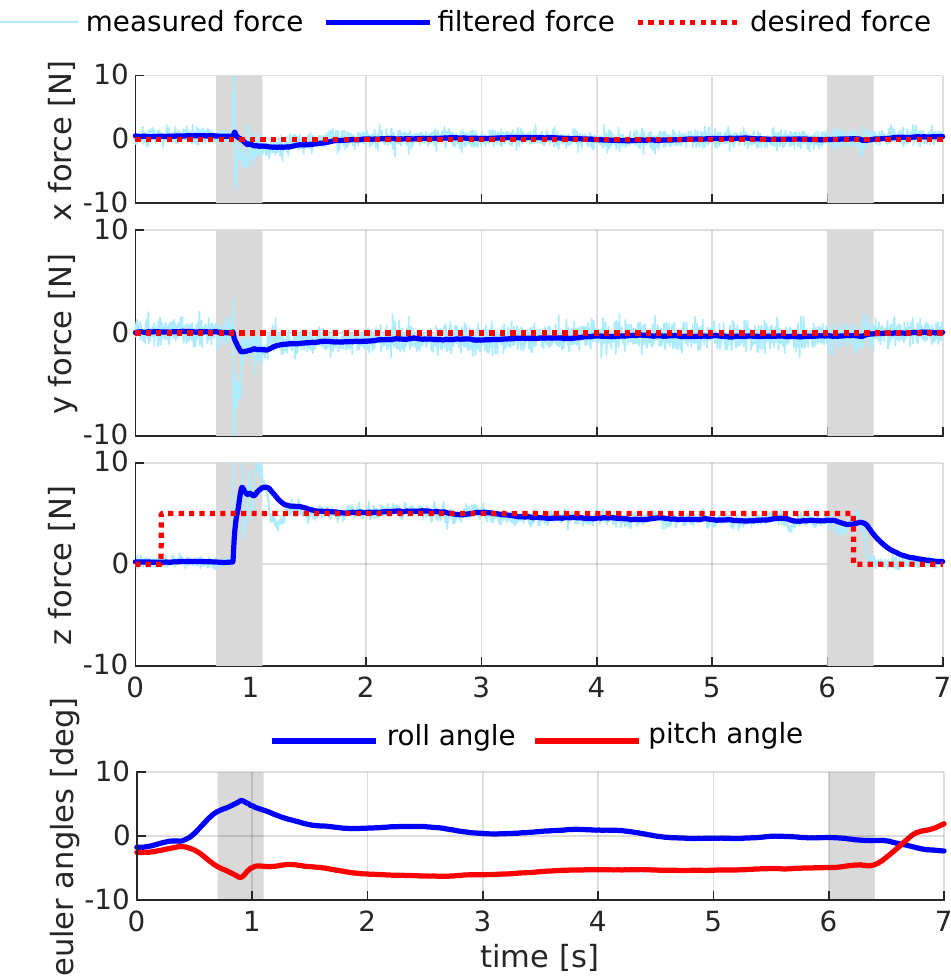}
  \end{subfigure}\quad
  \begin{subfigure}{8.5cm}
\centering\includegraphics[width=0.9\columnwidth]{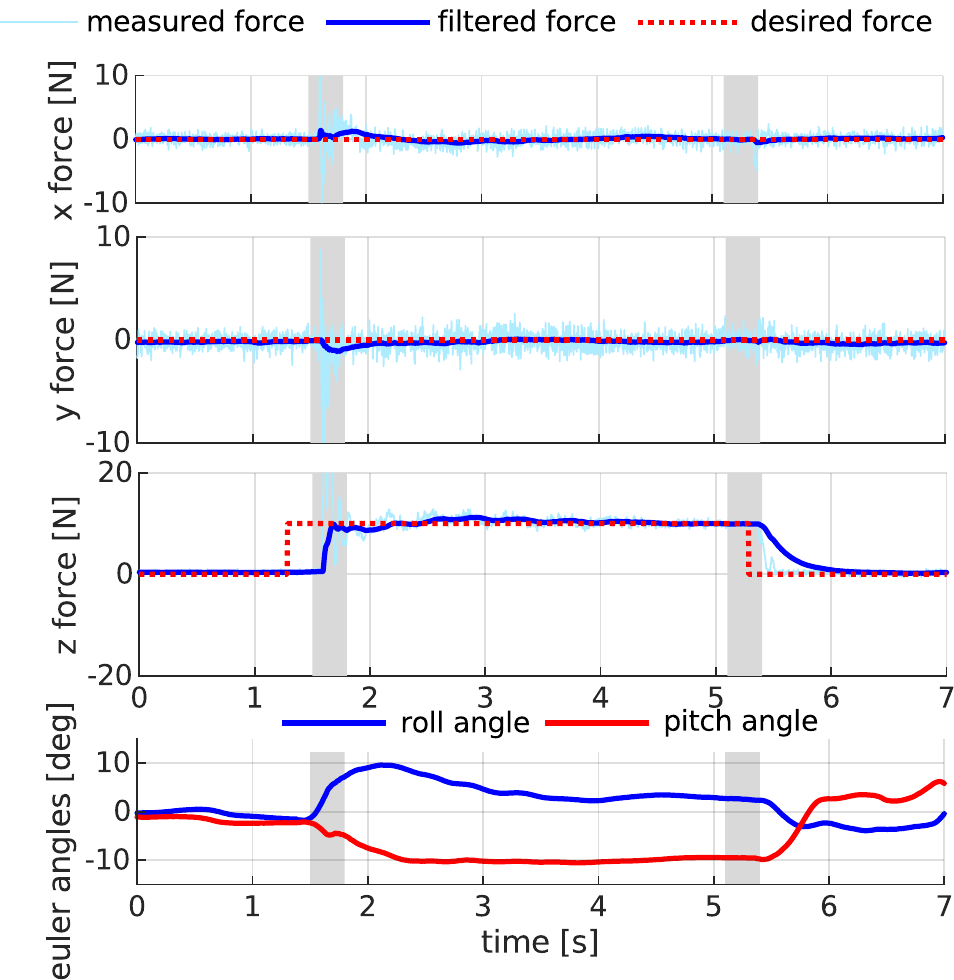}
  \end{subfigure}
  \caption{Evolution of end-effector forces and base orientation (roll and pitch angles) during a door pushing experiment. The shaded areas highlight the trajectories when the contact switch phases occur. In the left side figure, the commanded force in the end-effector normal direction is $5N$, while in the right side figure the commanded force is $10N$.}
  \label{fig:force_planning}
\end{figure*}

In Fig.~\ref{fig:force_planning}, the force tracking errors and the evolution of roll and pitch angles are shown, for two scenarios where the desired force is $5N$ and $10N$. The planned roll and pitch motion is such that the robot leans in order to exert a force and remain stable during the experiment. This generated behavior is very similar to that of a human trying to push a heavy door. Therefore, the base is more inclined when the commanded force is higher.
A demonstration of the door pushing experiment with a commanded force of $5N$ can be visualized in the attached video.

\section{Conclusion}

In this paper, we have addressed the control of mobile manipulators that need to dynamically balance, showing results on a newly designed ball-balancing robot with an arm. Due to its high non-linearity, non-minimum phase behaviour and inherent instability, this system can be considered as a technology demonstrator for all kinds of dynamically balancing mobile manipulators (e.g., humanoids, Segway-type systems and wheeled inverted pendulum with arms).

By defining the desired references as a soft constraint in the optimization, we have allowed the controller to plan for end-effector motion and balance over the same time horizon. By augmenting the action space of MPC with the end-effector forces, our robot has been shown to perform a door-opening task. We highlighted the different behaviors that can be obtained by appropriately choosing the cost function in the optimization and motivated the importance of using a whole-body approach for this kind of robots.

\subsection{Future Work}
Although this framework gives the flexibility to design the optimization problem based on the desired task, there are still more steps forward to take to improve our results. 
First, we intend to enhance the hardware platform to show more dynamic base and end-effector movements. Second, we have addressed the force planning problem considering only the robot and neglecting the world dynamics. 
Although it is not explicitly declared during the door opening experiment, by tracking different force references the optimization assigns an arbitrary impedance to the environment at its interaction port. 
This means that in order to be consistent with the physics of the environment, we should have added a set of state-input equality constraints to the MPC problem which relates the end-effector projected acceleration to the planned forces. 
However, we have not considered these constraints. Then, the question is: why does this controller work? 

The key to this issue is the replanning scheme of our controller. 
As long as we underestimate the impedance of the door (i.e., the reference forces are low enough), the door will move less than what our controller planned. Therefore, the contact remains closed and MPC can deal with this discrepancy by replanning from the current state. 
In our future work, we would like to look closer into this issue and improve the controller based on the estimation of the interaction port impedance. 

 \bibliographystyle{./myIEEEtran} 
 \bibliography{./IEEEabrv,IEEEexample,bibliography}

\end{document}

%% file: mathdef.tex
\newcommand{\vdelta}{\mbox{\boldmath $\delta$}}
\newcommand{\vep}{\mbox{\boldmath $\epsilon$}}

\newcommand{\vth}{\mbox{\boldmath $\theta$}}

\newcommand{\vlambda}{\mbox{\boldmath $\lambda$}}

\newcommand{\vnu}{\mbox{\boldmath $\nu$}}
\newcommand{\vxi}{\mbox{\boldmath $\xi$}}

\newcommand{\vtau}{\mbox{\boldmath $\tau$}}

\newcommand{\vom}{\mbox{\boldmath $\omega$}}

\newcommand{\vDe}{\bm \Delta}

\newcommand{\vPh}{\bm \Phi}

\newcommand{\ve}{\bm e}
\newcommand{\vf}{\bm f}

\newcommand{\vh}{\bm h}

\newcommand{\vp}{\bm p}
\newcommand{\vq}{\bm q}

\newcommand{\vu}{\bm u}
\newcommand{\vv}{\bm v}

\newcommand{\vx}{\bm x}

\newcommand{\vJ}{\bm J}
\newcommand{\vK}{\bm K}

\newcommand{\vM}{\bm M}

\newcommand{\vQ}{\bm Q}
\newcommand{\vR}{\bm R}
\newcommand{\vS}{\bm S}
\newcommand{\vT}{\bm T}





%% file: main.bbl
\begin{thebibliography}{10}
\providecommand{\url}[1]{#1}
\csname url@rmstyle\endcsname
\providecommand{\newblock}{\relax}
\providecommand{\bibinfo}[2]{#2}
\providecommand\BIBentrySTDinterwordspacing{\spaceskip=0pt\relax}
\providecommand\BIBentryALTinterwordstretchfactor{4}
\providecommand\BIBentryALTinterwordspacing{\spaceskip=\fontdimen2\font plus
\BIBentryALTinterwordstretchfactor\fontdimen3\font minus
  \fontdimen4\font\relax}
\providecommand\BIBforeignlanguage[2]{{%
\expandafter\ifx\csname l@#1\endcsname\relax
\typeout{** WARNING: IEEEtran.bst: No hyphenation pattern has been}%
\typeout{** loaded for the language `#1'. Using the pattern for}%
\typeout{** the default language instead.}%
\else
\language=\csname l@#1\endcsname
\fi
#2}}

\bibitem{satici2017intrinsic}
A.~C. Satici, A.~Donaire, and B.~Siciliano, ``Intrinsic dynamics and total
  energy-shaping control of the ballbot system,'' \emph{International Journal
  of Control}, vol.~90, no.~12, pp. 2734--2747, 2017.

\bibitem{nagarajan2012planning}
U.~Nagarajan, B.~Kim, and R.~Hollis, ``Planning in high-dimensional shape space
  for a single-wheeled balancing mobile robot with arms,'' in \emph{Robotics
  and Automation (ICRA), 2012 IEEE International Conference on}.\hskip 1em plus
  0.5em minus 0.4em\relax IEEE, 2012, pp. 130--135.

\bibitem{siciliano2010robotics}
B.~Siciliano, L.~Sciavicco, L.~Villani, and G.~Oriolo, \emph{Robotics:
  modelling, planning and control}.\hskip 1em plus 0.5em minus 0.4em\relax
  Springer Science \& Business Media, 2010.

\bibitem{bellicoso2019}
C.~D. Bellicoso, K.~Kramer, M.~St{\"a}uble, D.~Sako, F.~Jenelten, F.~Bjelonic,
  and M.~Hutter, ``Alma-articulated locomotion and manipulation for a
  torque-controllable robot,'' in \emph{International Conference on Robotics
  and Automation (ICRA 2019)}, 2019.

\bibitem{mason2018mpc}
S.~Mason, N.~Rotella, S.~Schaal, and L.~Righetti, ``An mpc walking framework
  with external contact forces,'' in \emph{2018 IEEE International Conference
  on Robotics and Automation (ICRA)}.\hskip 1em plus 0.5em minus 0.4em\relax
  IEEE, 2018, pp. 1785--1790.

\bibitem{zafar2016whole}
M.~Zafar and H.~I. Christensen, ``Whole body control of a wheeled inverted
  pendulum humanoid.'' in \emph{Humanoids}, 2016, pp. 89--94.

\bibitem{lauwers2006dynamically}
T.~B. Lauwers, G.~A. Kantor, and R.~L. Hollis, ``A dynamically stable
  single-wheeled mobile robot with inverse mouse-ball drive,'' in
  \emph{Robotics and Automation, 2006. ICRA 2006. Proceedings 2006 IEEE
  International Conference on}.\hskip 1em plus 0.5em minus 0.4em\relax IEEE,
  2006, pp. 2884--2889.

\bibitem{kumagai2008development}
M.~Kumagai and T.~Ochiai, ``Development of a robot balancing on a ball,'' in
  \emph{Control, Automation and Systems, 2008. ICCAS 2008. International
  Conference on}.\hskip 1em plus 0.5em minus 0.4em\relax IEEE, 2008, pp.
  433--438.

\bibitem{prieto2012monoball}
S.~S. Prieto, T.~A. Navarro, M.~G. Plaza, and O.~R. Polo, ``A monoball robot
  based on lego mindstorms [focus on education],'' \emph{IEEE Control Systems},
  vol.~32, no.~2, pp. 71--83, 2012.

\bibitem{fong2009design}
J.~Fong, S.~Uppill, and B.~Cazzolato, ``Design and build a ballbot,''
  \emph{Report, The University of Adelaide, Australia}, 2009.

\bibitem{shomin2015sit}
M.~Shomin, J.~Forlizzi, and R.~Hollis, ``Sit-to-stand assistance with a
  balancing mobile robot,'' in \emph{Robotics and Automation (ICRA), 2015 IEEE
  International Conference on}.\hskip 1em plus 0.5em minus 0.4em\relax IEEE,
  2015, pp. 3795--3800.

\bibitem{asgari2015manipulation}
P.~Asgari, P.~Zarafshan, and S.~A.~A. Moosavian, ``Manipulation control of an
  armed ballbot with stabilizer,'' \emph{Proceedings of the Institution of
  Mechanical Engineers, Part I: Journal of Systems and Control Engineering},
  vol. 229, no.~5, pp. 429--439, 2015.

\bibitem{fankhauser2010modeling}
P.~Fankhauser and C.~Gwerder, ``Modeling and control of a ballbot,'' {B.S.}
  thesis, Eidgen{\"o}ssische Technische Hochschule Z{\"u}rich, 2010.

\bibitem{Farshidian2014}
F.~{Farshidian}, M.~{Neunert}, and J.~{Buchli}, ``Learning of closed-loop
  motion control,'' in \emph{2014 IEEE/RSJ International Conference on
  Intelligent Robots and Systems}, 2014, pp. 1441--1446.

\bibitem{neunert2016fast}
M.~Neunert, C.~De~Crousaz, F.~Furrer, M.~Kamel, F.~Farshidian, R.~Siegwart, and
  J.~Buchli, ``Fast nonlinear model predictive control for unified trajectory
  optimization and tracking,'' in \emph{Robotics and Automation (ICRA), 2016
  IEEE International Conference on}.\hskip 1em plus 0.5em minus 0.4em\relax
  IEEE, 2016, pp. 1398--1404.

\bibitem{grandia2018frequency}
R.~Grandia, F.~Farshidian, A.~Dosovitskiy, R.~Ranftl, and M.~Hutter,
  ``Frequency-aware model predictive control,'' \emph{arXiv preprint
  arXiv:1809.04539}, 2018.

\bibitem{bostondynamics_2017}
BostonDynamics, ``Introducing handle,'' youtube. [Online]. Available:
  \url{https://www.youtube.com/watch?v=-7xvqQeoA8c}.

\bibitem{koenemann2015whole}
J.~Koenemann, A.~Del~Prete, Y.~Tassa, E.~Todorov, O.~Stasse, M.~Bennewitz, and
  N.~Mansard, ``Whole-body model-predictive control applied to the hrp-2
  humanoid,'' in \emph{2015 IEEE/RSJ International Conference on Intelligent
  Robots and Systems (IROS)}.\hskip 1em plus 0.5em minus 0.4em\relax IEEE,
  2015, pp. 3346--3351.

\bibitem{mirjalili2018whole}
R.~Mirjalili, A.~Yousefi-Korna, F.~A. Shirazi, A.~Nikkhah, F.~Nazemi, and
  M.~Khadiv, ``A whole-body model predictive control scheme including external
  contact forces and com height variations,'' in \emph{2018 IEEE-RAS 18th
  International Conference on Humanoid Robots (Humanoids)}.\hskip 1em plus
  0.5em minus 0.4em\relax IEEE, 2018, pp. 1--6.

\bibitem{zafar2018hierarchical}
M.~Zafar, S.~Hutchinson, and E.~A. Theodorou, ``Hierarchical optimization for
  whole-body control of wheeled inverted pendulum humanoids,'' \emph{arXiv
  preprint arXiv:1810.03074}, 2018.

\bibitem{bodie2016anypulator}
K.~Bodie, C.~D. Bellicoso, and M.~Hutter, ``Anypulator: Design and control of a
  safe robotic arm,'' in \emph{Intelligent Robots and Systems (IROS), 2016
  IEEE/RSJ International Conference on}.\hskip 1em plus 0.5em minus 0.4em\relax
  IEEE, 2016, pp. 1119--1125.

\bibitem{1905.06144}
R.~Grandia, F.~Farshidian, R.~Ranftl, and M.~Hutter, ``Feedback mpc for
  torque-controlled legged robots,'' 2019.

\bibitem{farshidian2017efficient}
F.~Farshidian, M.~Neunert, A.~W. Winkler, G.~Rey, and J.~Buchli, ``An efficient
  optimal planning and control framework for quadrupedal locomotion,'' in
  \emph{ICRA}.\hskip 1em plus 0.5em minus 0.4em\relax IEEE, 2017, pp. 93--100.

\bibitem{Farshidian2017MPC}
F.~Farshidian, E.~Jelavic, A.~Satapathy, M.~Giftthaler, and J.~Buchli,
  ``Real-time motion planning of legged robots: A model predictive control
  approach,'' in \emph{Humanoids}, 2017, pp. 577--584.

\bibitem{diehl2005}
M.~Diehl, H.~G. Bock, and J.~P. Schl{\"o}der, ``A real-time iteration scheme
  for nonlinear optimization in optimal feedback control,'' \emph{SIAM Journal
  on control and optimization}, vol.~43, no.~5, pp. 1714--1736, 2005.

\bibitem{giftthaler2017automatic}
M.~Giftthaler, M.~Neunert, M.~St{\"a}uble, M.~Frigerio, C.~Semini, and
  J.~Buchli, ``Automatic differentiation of rigid body dynamics for optimal
  control and estimation,'' \emph{Advanced Robotics}, vol.~31, no.~22, pp.
  1225--1237, 2017.

\bibitem{bloesch2018two}
M.~Bloesch, M.~Burri, H.~Sommer, R.~Siegwart, and M.~Hutter, ``The two-state
  implicit filter recursive estimation for mobile robots,'' \emph{IEEE Robotics
  and Automation Letters}, vol.~3, no.~1, pp. 573--580, 2018.

\end{thebibliography}
